\documentclass[10pt,conference]{IEEEtran}
\IEEEoverridecommandlockouts
\usepackage{cite}
\usepackage[switch]{lineno} 
\usepackage{amsmath,amssymb,amsfonts}
\usepackage{algorithmic}
\usepackage{graphicx}
\usepackage{textcomp}
\usepackage{subcaption}
\usepackage{xcolor}
\def\BibTeX{{\rm B\kern-.05em{\sc i\kern-.025em b}\kern-.08em
    T\kern-.1667em\lower.7ex\hbox{E}\kern-.125emX}}
\begin{document}

\title{Diagnosing Optimizer-Landscape Interaction via Representation in Subset Selection Multitasking: Evidence from Piano Fingering} 

\author{
    \IEEEauthorblockN{Ananda Phan Iman}
    \IEEEauthorblockA{\textit{Department of AI Convergence} \\
    \textit{Gwangju Inst. of Sci. \& Tech. (GIST)}\\
    South Korea \\
    anandaphan@gm.gist.ac.kr}
    \and
    \IEEEauthorblockN{Tingyang Wei}
    \IEEEauthorblockA{\textit{College of Computing and Data Science} \\
    \textit{Nanyang Technological University}\\
    Singapore \\
    tingyang001@e.ntu.edu.sg}
    \and
    \IEEEauthorblockN{Chang Wook Ahn}
    \IEEEauthorblockA{\textit{Department of AI Convergence} \\
    \textit{Gwangju Inst. of Sci. \& Tech. (GIST)}\\
    South Korea \\
    cwan@gist.ac.kr}    
}

\maketitle

\begin{abstract}
Continuous relaxations are commonly used as an encoding scheme for evolutionary multitasking optimization (EMTO), allowing heterogeneous tasks to be optimized within a unified search space. However, these continuous representations fundamentally reshape the induced fitness landscape, and their impact on optimization behavior and knowledge transfer remains underexplored. In this work, we bridge the gap by analyzing the interaction between the representation-induced fitness landscape and optimizer behavior through fitness landscape analysis. Using piano fingering estimation as a representative case of the subset selection problem, we introduce several encoding variants and investigate how their induced landscapes influence search dynamics and solution quality under EMTO. Our analysis indicates that transfer efficacy is strongly associated with changes in representation-induced landscape characteristics, rather than algorithmic factors alone, highlighting the representation design as the key factor of effective knowledge transfer in multitasking optimization.
\end{abstract}


\section{Introduction}
Evolutionary multitasking (EMTO)~\cite{Gupta2016,Gupta2017,wei2025} has been considered a powerful paradigm for solving multiple related optimization tasks concurrently by facilitating knowledge transfer through a unified search space. 
Despite its success in diverse applications~\cite{wei2022}, the effectiveness of EMTO in the discrete optimization is sensitive to the encoding scheme~\cite{Gupta2016}, which remains highly problem-dependent.
Specifically, enabling cross-task search requires constructing various continuous search spaces that represent heterogeneous tasks within a unified domain.
These efforts span classical combinatorial problems, such as job-shop scheduling problems~\cite{lirui2024}, vehicle routing problems~\cite{huang2024}, as well as more recent multitask evolutionary machine learning problems~\cite{Zhang2022, yang2025}.

Despite the existing works adopting random-key or specifically designed permutation representations~\cite{yuan2016,Feng2021} to relax multitask combinatorial optimization, the fundamental impact of such continuous representations on knowledge transfers remains largely unexplored. 
Since representation directly determines the geometry of the search space, it constrains what an optimizer can exploit~\cite{Ochoa2014LON}.
While discrete spaces lack smooth geometry, continuous relaxations provide \emph{soft} solutions with well-defined distances that enable more informative search dynamics.
Crucially, different relaxations on the same discrete solution can induce rather different fitness landscapes~\cite{Ichikawa2024CCRelaxation,bartunov2020continuous}.
Therefore, we are motivated to discuss how these representations affect search and knowledge transfer by investigating their landscape characteristics and how these features relate to multitasking performance.

\begin{figure}[t]
\centering
\includegraphics[width=0.9\columnwidth]{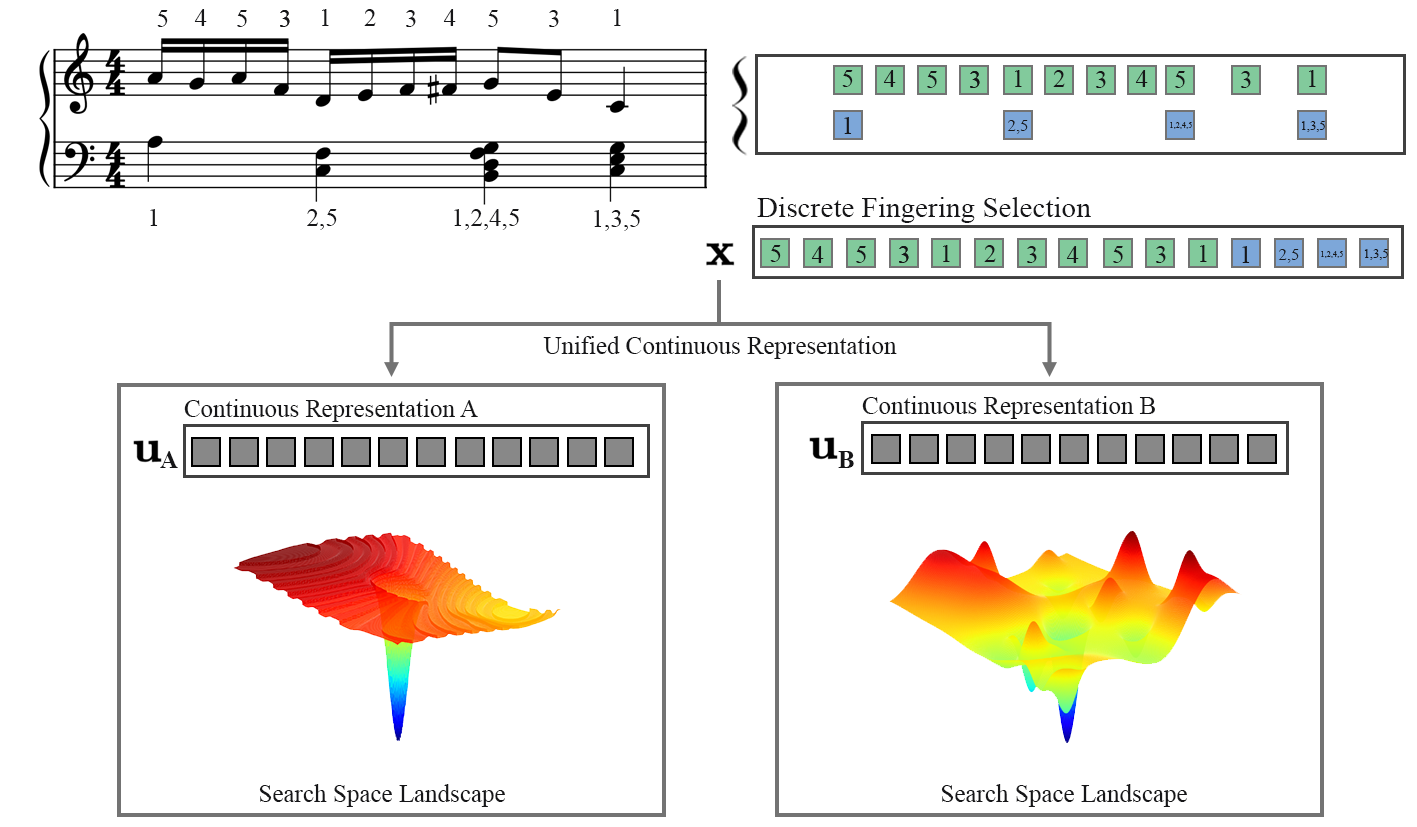} 
\caption{Illustration of how different unified continuous representations of the same discrete piano fingering selection $\mathbf{x}$ induce distinct search space. Top: A short piano passage with an example of fingering selection. Bottom: Two alternative unified continuous search spaces, within which optimization is performed; color indicates fitness values after decoding.}
\label{fig:introduction}
\end{figure}

To empirically investigate these representation-induced effects, we employ piano fingering estimation~\cite{iman2024model} as a representative instantiation of multitask subset selection~\cite{Qian2015SubsetSelection}.
In this problem, the objective is to determine an effective fingering sequence for a musical passage. 
At each time step $t$, a subset of $k_t$ fingers is selected from the five available fingers, where $k_t$ is determined by the number of notes played simultaneously.
Solving this problem within an EMTO framework thus requires translating the original discrete fingering selections into a unified continuous search space, followed by a decoding process that maps candidate solutions back to valid fingerings. 
However, this translation is not unique, as distinct encoding strategies can map the same discrete fingering selections into different continuous representations.
Such representation-dependent mappings directly shape the topology and geometry of the resulting search landscape.
Consequently, piano fingering provides a suitable environment for diagnosing how different representation-dependent landscapes influence search dynamics and knowledge transfer in evolutionary multitasking.

For illustration, the top part of Fig.~\ref{fig:introduction} shows a short musical passage in Western notation, where green and blue highlight notes played by the right and left hand, respectively. 
The lower part of the figure illustrates how different continuous representations of the same discrete fingering problem can form distinct landscape geometries.
Specifically, the same discrete fingering selection $\mathbf{x}$ admits multiple continuous representations ($\mathbf{u_A}$ and $\mathbf{u_B}$), each defining a different optimization landscape that corresponds to the same underlying discrete solution.
As shown in Fig. \ref{fig:introduction}, distinct representation-induced landscapes can possess rather different characteristics, potentially incurring diverse needs for search behavior or knowledge transfer behavior across tasks.
Therefore, this illustration highlights the role of continuous relaxation design in shaping the unified space explored by EMTO.
 
Despite the central role of continuous representation in defining the search landscape and facilitating transfer, to our knowledge, there is no systematic analysis of how different continuous relaxations of subset selection optimization behave under EMTO. 
Thus, it remains unclear how alternative continuous representations and their associated mapping affect solution quality, search dynamics and transfer performance.

In this work, we address this gap by analyzing how alternative continuous representations reshape the induced search landscape and affect optimizer behavior in subset selection multitasking. 
Using piano fingering estimation as a representative case, we empirically investigate the interaction between representation-induced search spaces and evolutionary dynamics to provide evidence-based insights into constructing continuous relaxations. 
While we focus on piano fingering in this paper, our findings offer broader implications for a wide class of subset selection problems, establishing principles for designing representations that facilitate effective search and reliable cross-task knowledge transfer under EMTO.

The relevant source code can be found in the link: https://github.com/anandaphan/pfmt-representation

\section{Related Work}
\subsection{Solution Representation in Evolutionary Multitasking}
Evolutionary Multitasking (EMTO), pioneered by \cite{Gupta2016}, aims to solve multiple tasks simultaneously through the implicit parallelism of evolutionary algorithms, formulated as:
\begin{equation}
\label{equation: emt}
\min f_q(\mathbf{x}_q),~~\text{s.t.} ~ \mathbf{x}_q \in \Omega_q,~~q \in \{ 1,\ldots,Q \}
\end{equation}
where $\Omega_q$ represents the decision space and $f_q$ denotes the objective function of the $q$-th task. 
While the original paradigm~\cite{Gupta2016} facilitates cross-task search within a unified search space, constructing an effective solution representation via \emph{explicit transfer}~\cite{Feng2019} has proven crucial for performance. 
Representation learning for transfer in EMTO has been explored via alignment-based approaches, including subspace alignment~\cite{Liang2022subspace} and manifold alignment~\cite{Chen2022Learning}.

However, these innovations primarily focus on continuous optimization. 
When $f_q(\cdot)$ in Eq.~\eqref{equation: emt} is a discrete optimization problem, representation learning becomes non-trivial and highly sensitive to problem-specific contexts~\cite{Gupta2016}. 
To facilitate implicit parallelism in discrete domains, EMTO traditionally employs random-key encoding~\cite{Bean1994GeneticAA, Gupta2016}. 
Alternative representations, such as the permutation-based approach proposed by \cite{yuan2016}, have demonstrated superior performance over random keys in specific domains like routing. 
Subsequent works have further refined these problem-dependent representations by incorporating domain heuristics~\cite{huang2023, Feng2021CVRP} and geometric distributions~\cite{Ray-access}.

In this work, we focus on a specific subclass of discrete optimization: subset selection~\cite{Qian2015SubsetSelection}. 
Formally, the subset selection problem seeks a binary decision vector $\mathbf{x} \in \{0, 1\}^n$ that minimizes $f$ under a constraint $k$:
\begin{equation}
\label{eq: subset}
\min_{\mathbf{x} \in \{0, 1\}^n} f(\mathbf{x}) \quad \text{s.t.} \quad \sum_{i=1}^{n} x_i \le k
\end{equation}
While this formulation is prevalent in multitask feature selection, the impact of solution representation in this context remains underexplored. 
For instance, some multitask optimizers operate directly on bitstrings~\cite{lin2024}, while others utilize continuous random keys in $[0,1]$ decoded via a simple threshold $\theta$~\cite{yang2025, Zhang2022}. 
Consequently, unlike continuous or permutation-based optimization, the impact of solution representation in subset selection multitasking remains unclear, particularly regarding how alternative continuous relaxations reshape the induced search landscape and influence transfer performance~\cite{Ichikawa2024CCRelaxation,bartunov2020continuous}.

\subsection{Solution Representation in Piano Fingering Estimation}
Piano fingering estimation provides an ideal case study for subset selection multitasking. At each time step, it instantiates an exact-$k$ subset selection problem: a specific subset of $k$ fingers must be selected to play $k$ notes simultaneously. 
Formally, this modifies the subset selection problem defined in Eq.~\eqref{eq: subset} by enforcing a cardinality constraint, $\sum_{i=1}^{n} x_i = k$. 

\begin{figure*}[t]
\centering
\begin{minipage}[t!]{0.45\textwidth}
    \centering
    \includegraphics[width=\columnwidth]{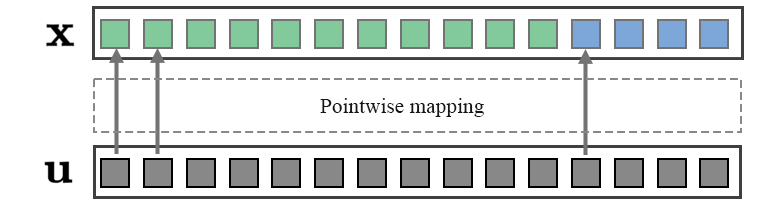}
    \label{fig:direct_representation}
    (1) R1: Direct (threshold-based)
    \par\vspace{2em}
    \includegraphics[width=\columnwidth]{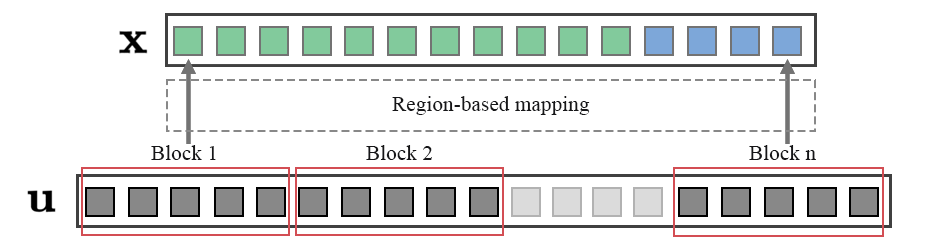}
    \label{fig:modular_representation}
    (2) R2--R3: Modular (block-structured)
\end{minipage}
\hfill
\begin{minipage}[t!]{0.5\textwidth}
    \centering
    \includegraphics[width=\columnwidth,keepaspectratio]{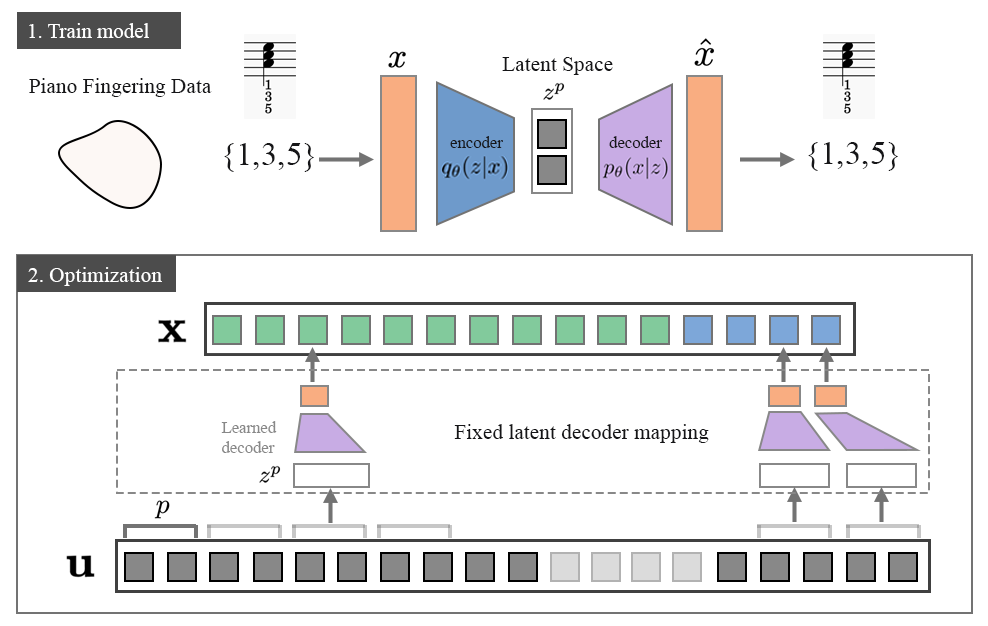}
    \label{fig:indirect_representation}
    (3) R4--R5: Indirect (latent-based)
\end{minipage}
\caption{Illustration of five representation variants(R1--R5) grouped by their underlying representation mechanisms. (1) R1 (Direct): each continuous variable is deterministically mapped to a discrete fingering selection. (2) R2--R3 (Modular): Fingering selections are controlled by blocks of continuous variables. (3) R4--R5 (Indirect): discrete fingerings are embedded in a low-dimensional latent space learned offline, with a fixed latent-to-discrete mapping during optimization.}
\label{fig:representation}
\end{figure*}

Traditional solutions to this problem include dynamic programming \cite{Kasimi2007}, tabu search \cite{Balliauw2015}, and variable neighborhood search \cite{Balliauw2017}, differential evolution \cite{De_Prisco2012fb}, hidden Markov models \cite{Nakamura2014, Nakamura2020}, and reinforcement learning \cite{iman2024model, gao2023generating, iman2025math13152485}. 
While these methods operate in discrete fingering spaces, we relax the problem into continuous representations to investigate the resulting search landscapes in our study.
This continuous relaxation approach draws inspiration from random-key encodings \cite{Bean1994GeneticAA, Gonalves2011BiasedRKGA, Rothlauf2002NetworkRK, Londe2025BRKGA}. It also relates to latent space optimization \cite{bartunov2020continuous, Lu2018SVAO, Luo2018NeuralOptimization}, where discrete structures are embedded into continuous manifolds. Unlike stochastic methods such as Gumbel-Softmax \cite{jang2017gumbelsoftmax} or Gumbel-Sinkhorn \cite{mena2018gumbelsinkhorn}. In contrast, our focus is on deterministic landscapes suitable for evolutionary search.

Piano fingering can be solved as multitasking by coupling multiple musical passages altogether. 
This allows us to diagnose how representation-dependent landscapes influence search dynamics and cross-task transfer efficiency under EMTO. To our knowledge, no study exists on multitasking music-relevant applications like piano fingering estimation. This absence highlights the novelty of our study as well.

\section{Methodology} \label{sec:methodology}
\subsection{Representation Variants}
In this section, we introduce several representation variants for mapping fingering subset selections into unified continuous search spaces.
Given a task $T_i$ with total musical time step $d=d_\text{right}+d_\text{left}$, a discrete fingering individual $\mathbf{x}=[x_1,x_2,\dots,x_d]$, where the search space for each gene is heterogeneous since each $x_j$ involves $k_j$ simultaneously played notes. 
Specifically, let $\mathcal{F}=\{1,2,3,4,5\}$ denote the set of finger indices. 
Then for the $j$-th gene with $k_j$ notes, the associated fingering search space $\mathcal{X}_j$ is given by the set of all $k_j$-subsets of $\mathcal{F}$:
\begin{equation}
    \mathcal{X}_{i,j} = [\mathcal{F}]^{k_j} = \{ A \subseteq \mathcal{F}  \; \mid \; |A| = k_j \}
\end{equation}

Moreover, different continuous representations map $\mathbf{x}$ into an $n$-dimensional unified search vector $\mathbf{u} \in [0,1]^n$, where $n=\sum_j dim(u_j)$. 
Although all representations operate over the same nominal domain, different continuous-discrete mappings induce distinct effective geometries. 
Accordingly, we treat each representation as a unified design consisting of continuous encoding together with the mapping operator that translates candidate solutions into discrete fingering selections for evaluations.
Based on these characteristics, we organize the representations into three types: direct, modular, and indirect.

Each representation variant is defined by a pair of $(\mathcal{U},g)$ where $\mathcal{U}\in [0,1]^n$ denotes the unified continuous search space and $g:\mathbf{u}\rightarrow\mathcal{X}_i$ is a deterministic mapping that translates continuous candidate solutions into discrete fingering selections for fitness evaluations. Given a candidate $\mathbf{u}\in\mathcal{U}$, the corresponding discrete fingering solution is $\mathbf{x}=g(\mathbf{u})$. The fitness function $f(x)$ is defined using the fingering difficulty level~\cite{iman2024model}.
\begin{equation} \label{eq:fitness_function}
\begin{aligned}
    &f(\mathbf{x}) = \frac{1}{d}\sum_{j=1}^{d}\sum_{l=1}^L{r_l(x_j})
\end{aligned}
\end{equation}
where $r_l(x_j)$ denotes the penalty incurred by the $l$-th fingering rule when applied to the $j$-th gene, reflecting the difficulty of executing the corresponding fingering. 
We adopt $L$ fingering rules defined in~\cite{iman2025math13152485} throughout this study.
Hence, different representations induce distinct fitness landscapes and search behaviors, even when the optimizer is unchanged.

\subsubsection{Direct Representation (R1)}
Direct representation employs a \emph{pointwise mapping}, in which each continuous gene $u_j$ corresponds to a single discrete fingering gene $x_j$, as illustrated in Fig.~\ref{fig:representation} (1). 
For the $j$-th time step requiring $k_j$ fingers, there are $M=\binom{5}{k_j}$ possible finger subsets. These subsets are ordered in a lexicographic order, and the unit interval $[0,1]$ is partitioned into $M$ contiguous subintervals of equal length, each associated with one subset. 
A continuous value $u_j$ is then mapped to the fingering subset associated with the unique interval $\left[\frac{m-1}{M}, \frac{m}{M}\right)$ containing $u_j$.

Due to hard partition boundaries, small perturbations of $u_j$ near these boundaries can induce abrupt changes in the resulting fingering. 
As a result, this representation exhibits minimal neutrality and high sensitivity, making it suitable as a baseline for analyzing the effects of a pointwise relaxation.

\subsubsection{Modular Representation (R2--R3)}
Motivated by the sensitivity of pointwise mappings to hard partition boundaries in R1, we consider modular representations that provide alternative continuous relaxations of the fingering selection problem. 
Modular representations employ a \emph{region-based mapping}, in which each fingering decision is controlled by a block of continuous variables rather than a single scalar, as illustrated in Fig.~\ref{fig:representation} (2).
As a result, modular representations naturally induce neutral regions in the fitness space.

\begin{figure}[t]
\centering
    \begin{subfigure}[t]{0.5\columnwidth}
        \centering
        \includegraphics[height=1.2in]{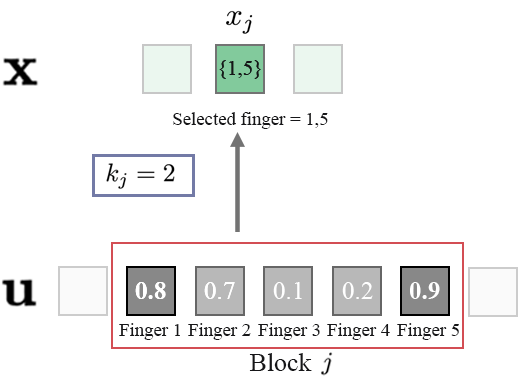}
        \caption{}
    \end{subfigure}%
    ~
    \begin{subfigure}[t]{0.5\columnwidth}
        \centering
        \includegraphics[height=1.2in]{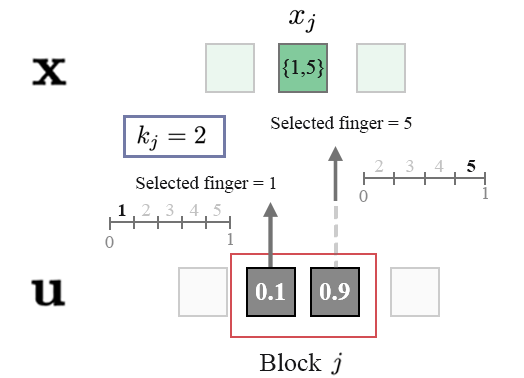}
        \caption{}
    \end{subfigure}
\caption{Example of region-based mapping for homogeneous and heterogeneous blocks. (a) homogeneous block with top-$k_j$ truncation mapping, (b) heterogeneous block with sequential selection without replacement.}
\label{fig:modular_homhem_example}
\end{figure}

In this work, we consider two types of block structures: \emph{homogeneous} and \emph{heterogeneous} blocks, which implement different selection rules illustrated in Fig.~\ref{fig:modular_homhem_example}.
\paragraph{Homogeneous blocks (R2)}
The homogeneous block representation follows a random-key-style encoding~\cite{Bean1994GeneticAA}, where each block $b_j\in[0,1]^5$ contains one value per finger. Given a required cardinality of $k_j$, mapping is performed at a block level. In R2, a \emph{truncation mapping}~\cite{Bean1994GeneticAA} selects the indices of top $k_j$ entries in $b_j$ and returns the corresponding fingers subset. 

\paragraph{Heterogeneous blocks (R3)}
In contrast, a heterogeneous block $b_j\in[0,1]^{k_j}$ contains one value per simultaneously played note. Mapping is performed by \emph{sequential selection without replacement} over the five fingers. Let $\mathcal{F}_0=(1,2,3,4,5)$ be the ordered list of available fingers. For $\ell=1, \dots,k_j$, $m_\ell=\min\{\lfloor b_{j,\ell}|\mathcal{F}_{\ell-1}|\rfloor,\ |\mathcal{F}_{\ell-1}|-1\}$, select the $m_\ell$-th element of $\mathcal{F}_{\ell-1}$ and remove it to obtain $\mathcal{F}_\ell$. The selected fingers are then sorted to form an unordered $k_j$-subset and thus always $k_j$ distinct fingers.

\subsubsection{Indirect (Latent) Representation (R4--R5)}
The indirect representation introduces a low-dimensional continuous encoding of discrete fingering configurations using real fingering data from the PIG Dataset \cite{Nakamura2020}. 
Each discrete fingering subset $x_j$ is represented by a compact latent vector $\mathbf{z}_j \in \mathbb{R}^p$, enabling search to be conducted in a semantically structured latent space rather than the discrete domain.

Specifically, we employ a variational autoencoder (VAE)~\cite{Kingma2014VAE}, where the encoder $q_\phi(\mathbf{z}_j|x_j)$ maps $x_j$ to $\mathbf{z}_j\in \mathbb{R}^p$, and a decoder $p_\theta(x_j|\mathbf{z}_j)$ reconstructs the corresponding fingering. The VAE model is trained offline by minimizing the standard evidence lower bound (ELBO).
After training, the encoder is discarded, and the decoder is fixed and used as a mapping during optimization.

To obtain a discrete fingering solution, $\mathbf{u}$ is divided into $d$ segments $\mathbf{u}_j\in [0,1]^p$ which are transformed into latent space via a bijective transformation $h(\cdot)$, and subsequently mapped using the fixed VAE decoder $g_\theta(\cdot)$:
\begin{equation}
\begin{aligned}
    h&:[0,1]^p \rightarrow \mathbb{R}^p, \qquad &z_j=h(u_j)\\
    g_\theta&:\mathbb{R}^p \rightarrow \mathcal{X}_{i,j}, \qquad &x_j=g_\theta(z_j)\\
\end{aligned}
\end{equation}
The indirect representation is illustrated in Fig.~\ref{fig:representation} (3). We consider two latent dimensionalities, $p=2$ (R4) and $p=1$ (R5), to examine the effect of latent compression in the induced search landscape. By construction, nearby latent points tend to decode to identical or similar fingering selections, inducing a landscape informed by real pianistic data. 

Together, the three types of representations span a spectrum from one-to-one deterministic to data-driven, each inducing a distinct continuous search space and interacting differently with evolutionary search and knowledge transfer under EMTO.

\begin{table}
    \centering
    \caption{Representation Variant Summary}
    \begin{tabular}{llp{5cm}}
        \hline
        ID & Type & Mapping \\
        \hline
        R1 & Direct & Partition-based pointwise mapping\\
        R2 & Modular (hom.) & top-$k_j$ truncation mapping\\
        R3 & Modular (het.) & Sequential selection without replacement\\
        R4 & Indirect & Fixed VAE decoder, $p=2$ \\
        R5 & Indirect & Fixed VAE decoder, $p=1$ \\
        \hline
    \end{tabular}
    \label{tab:representation-summary}
\end{table}

\subsection{Fitness Landscape Analysis}\label{sec:method-fla}
To characterize how continuous representations reshape the search space explored by EMTO, we analyze the representation-induced landscapes using random walks.
Although a wide range of landscape measures exists~\cite{Zou2022fla}, we focus on ruggedness and neutrality as they directly capture fitness variation and invariance under local perturbations, and are widely used in fitness landscape analysis (FLA) \cite{Weinberger1990Autocorrelation,Barnett1998RuggednessNeutrality,vanneschi2007neutrality}.

For each representation with unified dimension $n$, we perform random walks on $[0,1]^n$ using normalized Gaussian steps, 
$x_{t+1}=\mathcal{B}(x_t+\eta z_t/\|z_t\|_2)$ with $z_t\sim\mathcal{N}(0,I)$, step size $\eta$, and reflection bounds. Each visited point is decoded into a fingering solution via $g(\cdot)$ and evaluated to obtain the fitness trace $F_t=f(g(x_t))$.

\paragraph{Ruggedness}
Ruggedness reflects how rapidly fitness changes under small perturbations~\cite{Zou2022fla}. We quantify ruggedness via random walk autocorrelation of $\{F_t\}$, using the lag-1 autocorrelation $\rho(1)$ and autocorrelation length $\tau$ \cite{Weinberger1990Autocorrelation,Stadler2002}. Higher $\rho(1)$ indicates smoother landscapes.

\paragraph{Neutrality}
Neutrality captures plateau-like behavior where local perturbations do not change the mapped objective~\cite{Barnett1998RuggednessNeutrality,Zou2022fla}. We compute neutrality $\nu$ by measuring $\varepsilon$-neutrality, $\nu_\varepsilon=\frac{1}{W}\sum_{t=0}^{W-1}\mathbb{I}(|F_{t+1}-F_t|\le \varepsilon)$, where W is the total visited point. We set $\varepsilon$ per representation as $\varepsilon=\gamma(\max\Delta F-\min\Delta F)$ with $\gamma=0.001$, where $\Delta F=F_{t+1}-F_t$.

Together, ruggedness and neutrality provide a principled basis for analyzing the fitness landscapes induced by each representation.

\section{Experimental Study}\label{sec:experiment}

\subsection{Experiment Setting}
We construct eight task pairs from PIG dataset \cite{Nakamura2020} with controlled musical relatedness. Relatedness is defined by key-signature agreement and tonal relationship (same key, relative key, different keys, or unrelated), inducing different levels of structural overlap between tasks (complete/partial/none). Cases 1-4 use major keys and cases 5-8 use minor keys, with cases 3 and 7 involving relative key pairings. 
Thus, for each task pair $T=\{T_1, T_2\}$, we set population size per task to $N=50$ and $10^5$ evaluations per task, with fitness computed using Eq.~\eqref{eq:fitness_function}.

To assess representation effects across optimizers, we evaluate the recent EMTO algorithms spanning the three canonical operator families: GA, DE, and ES. 
Specifically, \emph{GA-based} methods include AT-MFEA \cite{Xue2022AT_MFEA}, MFEA-AKT \cite{Zhou2021MFEA_AKT}, MFEA-DGD \cite{Liu2024MFEA_DGD}, MFMP \cite{Li2020MFMP}, and MTEA-AD \cite{Wang2021MTEA_AD}. \emph{DE-based} methods include BLKT-DE \cite{Jiang2024BLKT_DE} and  MKTDE \cite{Li2022MKTDE}. Lastly, \emph{ES-based} methods include MTES-KG \cite{Li2024MTES_KG}. These algorithms span both multifactorial and multipopulation paradigms, and we use the MTO Platform~\cite{Li2023MToP} implementation and adapted our piano fingering tasks to their evaluation interface without modifying the algorithmic operators or default hyperparameters.

We evaluate five representations summarized in Table~\ref{tab:representation-summary}. For the VAE-based representations (R4-R5), we use a VAE with two hidden MLP layers (64 and 32 units, LeakyReLU), trained for 40 epochs (batch size 512) with Adam (lr $=5\times10^{-3}$). The latent mapping $h(\cdot)$ linearly rescales $u_j\in[0,1]^p$ to $z_j\in[-3,3]^p$ via $z_j=6u_j-3$.

For each representation-algorithm pair, 32 independent runs are conducted to characterize both optimization performance and behavior. Performance is evaluated using the average of the best objective value on each task and the aggregate metrics. 
To implement the fitness landscape analysis described in Section \ref{sec:method-fla}, for each representation, we perform $50$ independent random walks in the unified space, each with $5000$ visited points, using a normalized Gaussian neighborhood ($\eta=0.05$). Fitness values along the walks are used to compute the lag-1 autocorrelation $\rho(1)$, autocorrelation length $\tau$, and neutrality $\nu$, which are then correlated with EMTO performance.

\begin{figure}[tbh]
\centering
\includegraphics[width=0.88\columnwidth]{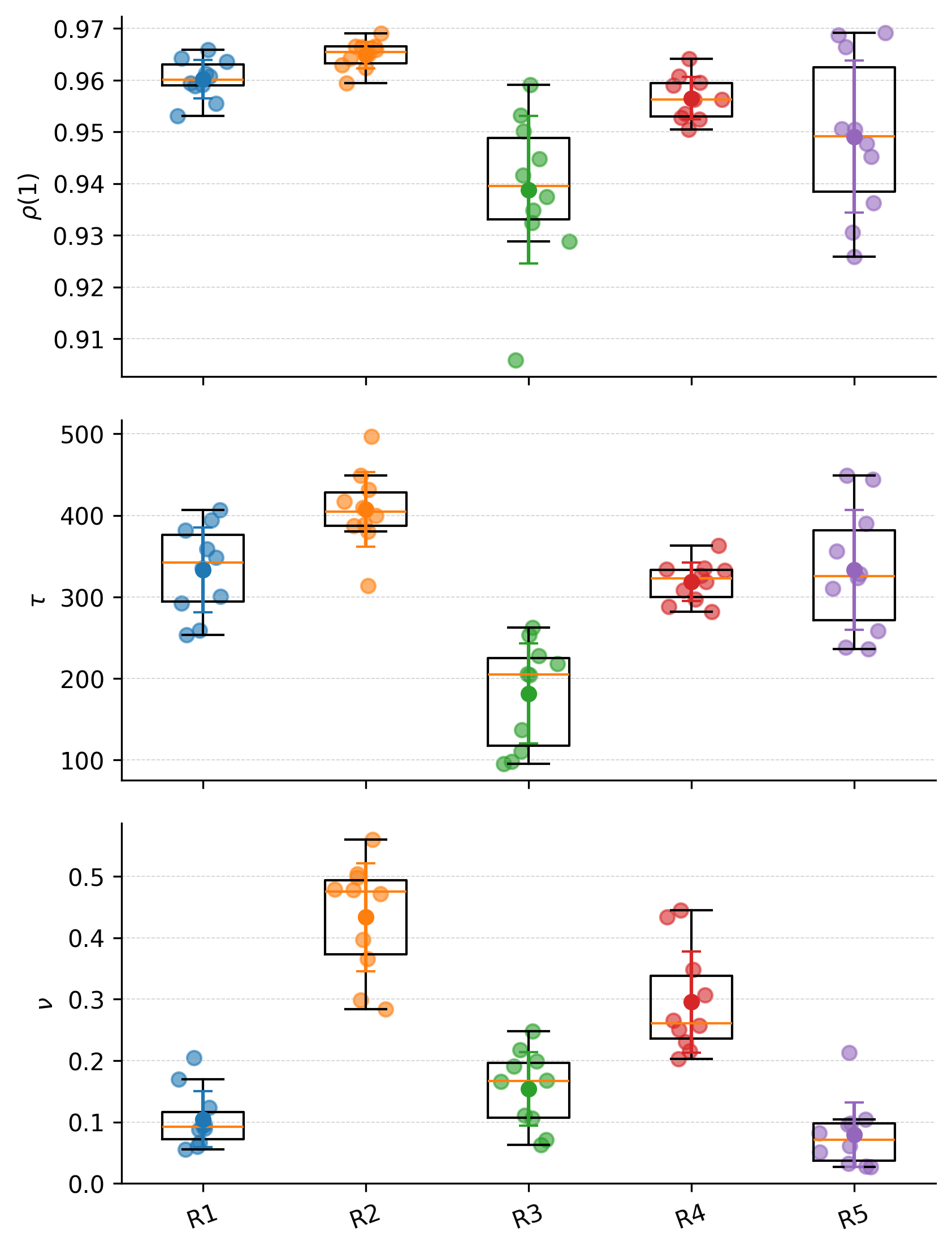}
\caption{Distribution of representation-induced landscape features of R1-R5 across tasks, including lag-1 autocorrelation $\rho(1)$, autocorrelation length $\tau$, and neutrality $\nu$.}
\label{fig:landscape_analysis}
\end{figure}

\begin{table*}
    \centering
    \caption{Median percent transfer effect $\Delta\ (\%)$ across 8 tasks, $\Delta\ (\%)=(f^\text{STO}-f^\text{MTO})/f^\text{STO} \times100$ (minimization, matched STO per optimizer family). $\Delta<-100$ indicates $f^\text{MTO}>2f^\text{STO}$. Bold indicates the largest $\Delta$ (best transfer effect) per family.}
    \begin{tabular}{|c||c|rrrrr||c|rr||c|r|}
        \hline
        &&\multicolumn{5}{c||}{$\Delta\ (\%)$}& & \multicolumn{2}{c||}{$\Delta\ (\%)$} && $\Delta\ (\%)$\\
        \cline{3-7} \cline{9-10} \cline{12-12}
        Rep. & GA & AT-MFEA & MFEA-AKT & MFEA-DGD & MFMP & MTEA-AD & DE & BLKT-DE & MKTDE & CMA-ES & MTES-KG\\
        \hline
        R1 &---& $+0.78\%$&   $-6.54\%$ & $-136.34\%$&	$-9.80\%$ & $+1.06\%$	&---&	$+8.11\%$ &$-3.82\%$     &---&	    $-26.27\%$ \\
        R2 &---& $+7.01\%$&   $-26.35\%$ & $-106.73\%$&	$-12.35\%$ & $+7.26\%$	&---&	$+11.56\%$ &	$+2.74\%$ &---&	    $-106.03\%$ \\
        R3 &---& $+5.62\%$&   $-15.02\%$ & $-98.69\%$&	$-56.29\%$ & $\mathbf{+9.26\%}$ &---&   $\mathbf{+27.27\%}$ &	$+13.70\%$ &---&	$-43.83\%$ \\
        R4 &---& $+3.50\%$&   $-13.11\%$ & $-101.10\%$&	$-6.34\%$ & $+4.82\%$	&---&	$+5.15\%$ &	$+0.89\%$  &---&	$-20.93\%$ \\
        R5 &---& $+0.23\%$&   $-7.64\%$ & $-116.24\%$&	$-11.27\%$ & $+1.71\%$	&---&	$+5.69\%$ &	$-9.33\%$ &---& 	$-6.06\%$  \\
        \hline
    \end{tabular}
    \label{tab:improvement}
\end{table*}

\begin{figure*}[thb]
\centering
\includegraphics[width=\linewidth]{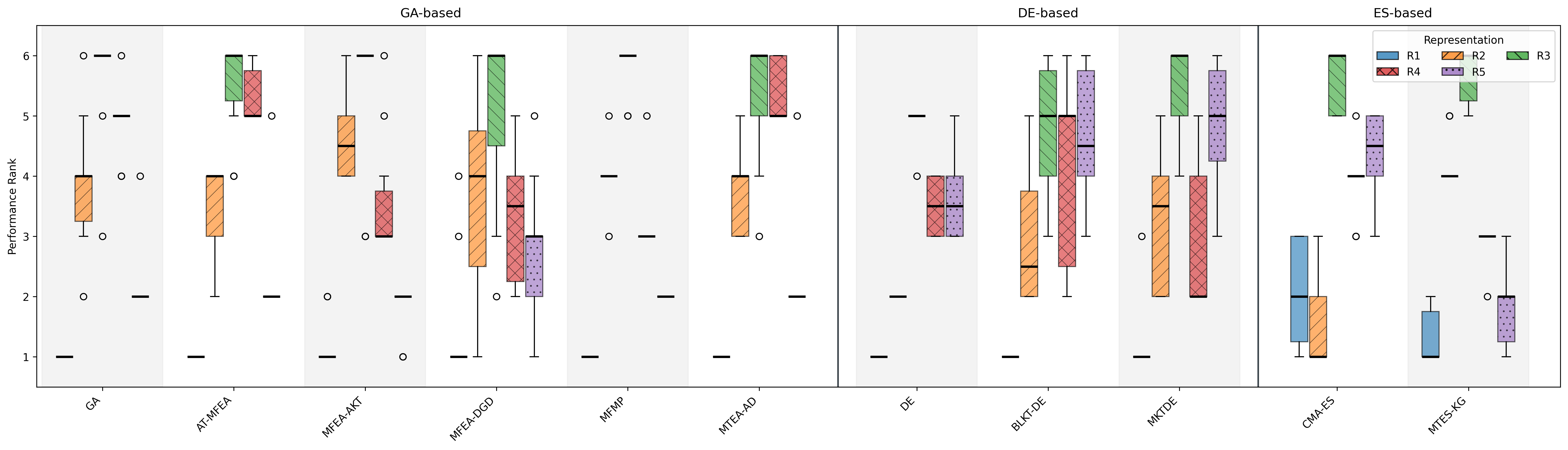} 
\caption{Performance rank distributions across representations and optimizer. Boxplots show the median and interquartile range (IQR) across all tasks and runs. 
Collapsed boxes indicate zero IQR (stable central performance). Optimizers are grouped by their family; lower ranks indicate better performance.}
\label{fig:multitasking_performance}
\end{figure*}

\subsection{Landscape Characterization Across Representations}
Fig.~\ref{fig:landscape_analysis} summarizes the distribution of representation-induced landscape features. All landscapes (R1-R5) exhibit smooth landscapes with high lag-1 autocorrelation ($\rho(1)>0.90$) and long correlation length ($\tau>95$), indicating coherent neighborhoods in the unified space. 
The main differentiator is neutrality $\nu$, which controls the prevalence of plateaus and thus the balance between directed search and neutral drift.

Although all landscapes are smooth, correlation persistence differs across representations. 
R2 exhibits the longest autocorrelation lengths, forming wide and coherent basins that support efficient long-range exploitation.  In contrast, R3 shows the shortest $\tau$ despite being locally smooth, resulting in faster correlation decay. R1, R4, and R5 lie between these extremes.

Neutrality separates the representations most clearly: R2 is dominated by extended neutral plateaus, making it favorable for drift-based and covariance-based optimizers. This suggests that homogeneous blocks in R2 induce plateau-dominated landscapes.
Conversely, R1, R3, and R5 exhibit low neutrality, resulting in more decision-forced and slope-driven landscapes. Furthermore, lowering the latent size from R4 to R5 leads to a decrease in neutrality.

\subsection{Representation Effect on Multitasking Performance}
Table~\ref{tab:improvement} summarizes the relative transfer effects and Fig.~\ref{fig:multitasking_performance} shows the performance rank distributions across representations and algorithms. Overall, multitasking performance depends strongly on both representation and optimizer, and positive transfer is not guaranteed. 

Particularly, under AT-MFEA, MTEA-AD, BLKT-DE, and MKTDE, R2 achieved up to $+7\%$ (GA-based) and $+11$-$12\%$ (DE-based). Moreover, R3 shows particularly strong gain for DE-based multitasking, achieving improvements of $+27.27\%$ in BLKT-DE and $+13.70\%$ under MKTDE.
In contrast, R1 and R5 yield only small positive gains for AT-MFEA and MTEA-AD, with the improvements noticeably smaller than those observed in R2-R3. Both also experience severe performance degradation under MKTDE, while the others show improvements. This suggests that R1 and R5 provide limited transferable structure and only weakly support multitasking. Furthermore, MFEA-AKT, MFMP, and MTES-KG consistently underperform across all representations with degradation ranging from $-6\%$ to $-106\%$, indicating systematic negative transfer. 

Beyond the performance improvement, Fig.~\ref{fig:multitasking_performance} reveals an important difference in performance consistency and dominance through rank distributions.
Notably, R1 frequently attains the best rank (rank 1) across multiple optimizers and tasks, reflecting its strong task efficiency and aggressive search behavior.  However, this dominance is not uniformly preserved across multitasking strategies. Particularly, under the ES-based multitasking optimizer MTES-KG, the strong performance of R2 observed under CMA-ES deteriorates substantially, with ranks dropping to 4-5 and performance degrading by up to $-106\%$, consistent with a breakdown in transfer effectiveness when optimizer-landscape compatibility is violated.

To assess whether the observed differences indicate a globally dominant representation, a Friedman Test was conducted across all representations and the multitasking optimizer. The results indicate a significant global performance separation among representations. Furthermore, pairwise Wilcoxon signed-rank tests ($\alpha=0.05$) show that R1 achieves statistically superior performance compared to the remaining representations when aggregated across all settings. Taken together, the variability in improvement and rank stability across optimizers suggests that these performance differences are driven not solely by representation choice, but also by optimizer–landscape interaction. 

\begin{figure}[t]
\centering
\includegraphics[width=\columnwidth]{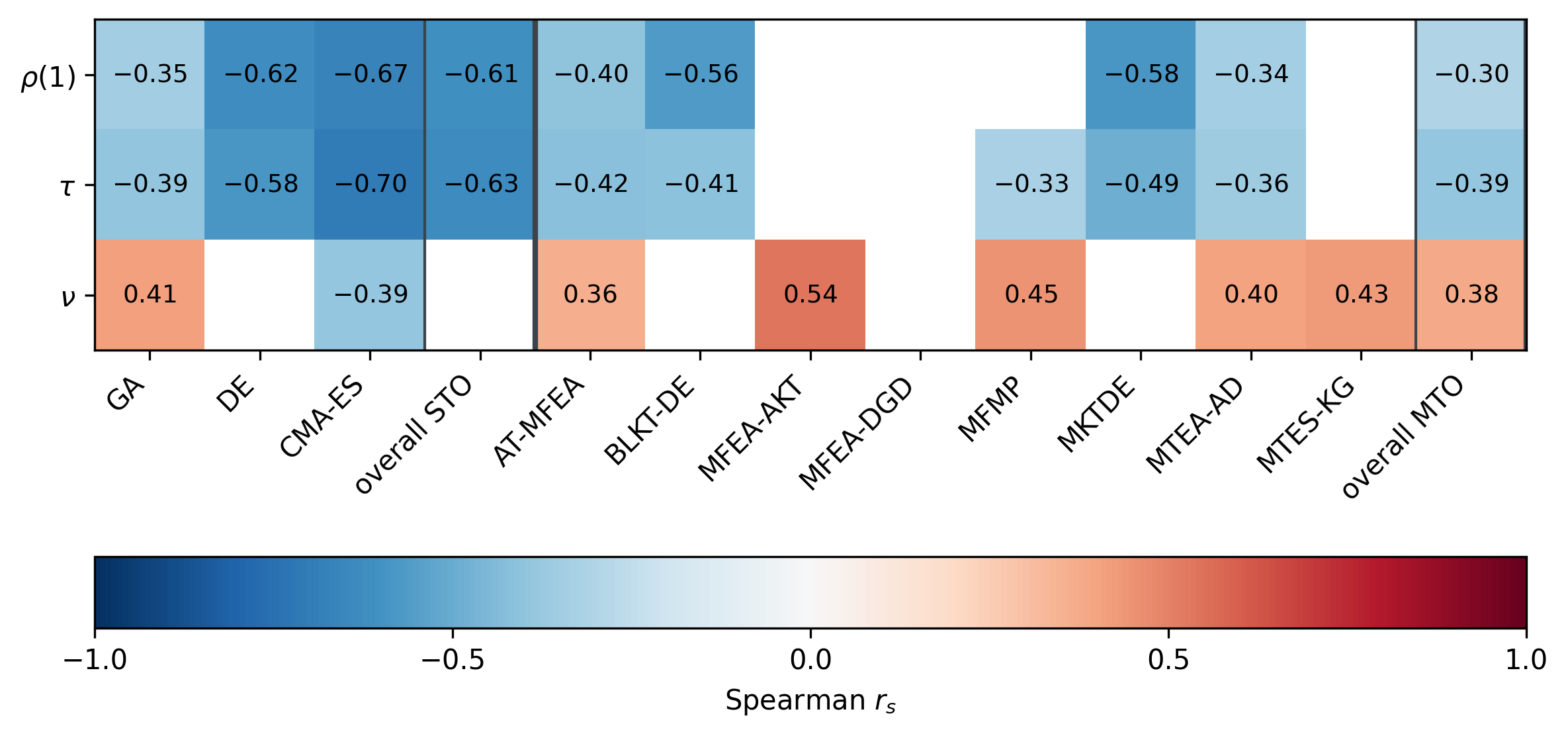}
\caption{Heatmap of Spearman correlation ($r_s$) between algorithm performance and landscape, including lag-1 autocorrelation $\rho(1)$, autocorrelation length $\tau$, and neutrality $\nu$. Columns grouped by optimizer (STO and MTO), and only statistically significant with Holm-corrected ($\alpha=0.05$) are shown; others are blanks.}
\label{fig:correlation}
\end{figure}

\subsection{Optimizer-Landscape Interaction in STO and MTO}
The performance trends observed in the previous subsection indicate that the same representation-optimizer pair can lead to markedly different performance outcomes once transfer is introduced.
This observation suggests that multitasking optimization (MTO) alters the relationship between optimizer dynamics and the representation-induced landscape.
To better understand this behavior, we analyze this relationship via Spearman correlations between landscape features and performance under single task optimization (STO) and MTO as shown in Fig.~\ref{fig:correlation}.

Under STO, ruggedness-related features $\rho(1)$ and $\tau$ dominate across optimizers, while neutrality plays a secondary and optimizer-dependent role. Higher neutrality tends to degrade GA performance, strongly benefits CMA-ES, and is largely irrelevant for DE. This behavior is consistent with the predominantly local nature of STO, where progress is driven by exploitation of local structure.

Under MTO, neutrality becomes substantially more predictive and ruggedness remains relevant, indicating that successful transfer requires representations that can preserve useful partial structures under cross-task variation and a coherent local neighborhood in the unified space. Thus, MTO performance is influenced by \emph{both} ruggedness and neutrality.

In our MTO setting, mapping-induced neutral plateaus can buffer cross-task transfer, allowing offspring to inherit transferable components without immediate degradation, thereby mitigating negative transfer. Although all representations yield relatively smooth traces (high $\rho(1)$), smoothness can still limit post-transfer refinement: representations with stronger local coherence enable faster exploitation after mixing. Hence, effective MTO requires representations that support safe information exchange and maintain usable local structure for exploitation, helping explain why improvements arise only when both properties are present.

At the algorithmic level, MTO reveals a clear difference in how different optimizers respond to representation-induced landscape features. 
AT-MFEA and MTEA-AD exhibit the strongest coupling to landscape structure, with performance significantly associated with both ruggedness-related metrics and neutrality, suggesting they benefit from neutral buffers for exchange while still relying on coherent local neighborhoods for refinement.
In contrast, methods such as MFEA-AKT and MTES-KG are primarily associated with neutrality, with smoothness metrics showing little explanatory power. Their performance improves as neutrality decreases, consistent with smaller degradation observed for R1 and R5 under these optimizers. MFMP is sensitive to $\tau$ and $\nu$ but not $\rho(1)$, suggesting it responds to correlation length and neutrality rather than purely to lag-1 continuity.
DE-based multitasking optimizers still remain strongly associated with smoothness-related metrics and limited sensitivity to neutrality, indicating that DE operators continue to rely on local continuity even under multitasking. On the other hand, MFEA-DGD shows no consistent associations with any landscape metric, coinciding with its consistently poor performance and suggesting limited ability to exploit induced landscape structure.

Overall, AT-MFEA and MTEA-AD are the most structure-aligned optimizers, as they are the only methods whose performance ranks are significantly associated with both smoothness and neutrality-related landscape metrics.

\subsection{Discussion}
This study shows that MTO alters the relationship between optimizer dynamics and representation-induced landscape structure. While STO is largely explained by the ruggedness-related features, MTO increases the importance of neutrality together with ruggedness. This suggests that effective multitasking requires both neutral regions and coherent local neighborhoods. This distinction clarifies why representation-optimizer pairs that perform well in isolation may fail or even degrade once multitasking is introduced. Practically, representation choice should be optimizer-aware, since transfer changes which landscape properties are predictive.

Our findings further suggest that optimizer design embodies implicit assumptions about the geometry of the induced landscape in both single and multitasking settings. When transfer is introduced, cross-task recombination induces larger and less local perturbations in the unified space, which can invalidate these assumptions and lead to negative or catastrophic transfer. Consequently, no representation can be universally optimal across optimizers. Effective multitasking emerges only when the transfer mechanism is compatible with the induced landscape structure. These results indicate that representation should be treated as a first-order design factor in subset selection multitasking optimization, as it fundamentally shapes the landscape properties exploited during transfer. 

However, several limitations should be noted. First, the analysis is conducted on subset selection problems. Further validation is thus required to assess its generality across other problem domains. Second, the landscape-performance relationship is established through correlation and does not imply direct causality. Third, although multiple optimizer families are examined, we do not exhaustively cover all transfer mechanisms. Finally, extending these findings to adaptive or co-evolving representations remains an important direction for future work.

\section{Conclusion} \label{sec:conclusion}
This work presents a systematic analysis of how representation-induced landscape structure interacts with the optimizer dynamics in evolutionary multitasking optimization. Using piano fingering as the representative case of subset selection multitasking, we examine the performance of various continuous representations under both single-task and multitasking optimization. 
Our results show that multitasking changes which landscape properties influence optimization behavior. While ruggedness dominates STO, MTO performance depends on both neutrality and ruggedness. These findings highlight the importance of optimizer-landscape compatibility and demonstrate that representation plays a central role in multitasking optimization.

\section*{Acknowledgments}
This work was supported by Institute of Information \& communications Technology Planning \& Evaluation (IITP) grant funded by the Korea government (MSIT) (No.2019-0-01842, Artificial Intelligence Graduate School Program (GIST)).

\bibliographystyle{ieeetr}
\bibliography{pianofingering}

\end{document}